\documentclass[12pt,twoside]{article}

\usepackage{amsmath,amssymb}
  \newdimen\paravsp  \paravsp=1.3ex
\newtheorem{theorem}{Theorem}
\newtheorem{lemma}[theorem]{Lemma}
\newtheorem{definition}[theorem]{Definition}
\newtheorem{example}[theorem]{Example}
\newtheorem{proposition}[theorem]{Proposition}
\newtheorem{conjecture}[theorem]{Conjecture}
\newtheorem{remark}[theorem]{Remark}
\newenvironment{keywords}{\centerline{\bf\small
Keywords}\begin{quote}\small}{\par\end{quote}\vskip 1ex}
\newenvironment{proof}{\paradot{Proof}}{}
\def\beq{\begin{equation}}    \def\eeq{\end{equation}}
\def\beqn{\begin{displaymath}}\def\eeqn{\end{displaymath}}
\def\bqa{\begin{eqnarray}}    \def\eqa{\end{eqnarray}}
\def\bqan{\begin{eqnarray*}}  \def\eqan{\end{eqnarray*}}
\newdimen\paravsp  \paravsp=1.3ex
\def\paradot#1{\vspace{\paravsp plus 0.5\paravsp minus 0.5\paravsp}\noindent{\bf\boldmath{#1.}}}

\def\eps{\varepsilon}
\def\epstr{\epsilon}                    

\def\qed{\hspace*{\fill}\rule{1.4ex}{1.4ex}$\quad$}
\def\eoe{\hspace*{\fill} $\diamondsuit\quad$} 
\def\eor{\hspace*{\fill} $\diamond\quad$} 

\def\S{{\cal S}}
\def\X{{\cal X}}
\def\Y{{\cal Y}}
\def\EE{I\!\!E} 
\def\t{\theta}
\def\Cost{\mbox{\sl Cost}}
\def\ICost{\mbox{\sl ICost}}
\def\OCost{\mbox{\sl OCost}}

\begin{document}

\title{
\vskip 2mm\bf\Large\hrule height5pt \vskip 4mm
Consistency of Feature Markov Processes
\vskip 4mm \hrule height2pt}
\author{{\bf Marcus Hutter} and {\bf Peter Sunehag} \\[3mm]
\normalsize RSISE$\,$@$\,$ANU and SML$\,$@$\,$NICTA \\
\normalsize Canberra, ACT, 0200, Australia \\
\texttt{\small\{Peter.Sunehag,Marcus.Hutter\}@anu.edu.au}
}
\date{12 July 2010}

\maketitle

\begin{abstract}
We are studying long term sequence prediction (forecasting). We
approach this by investigating criteria for choosing a compact
useful state representation. The state is supposed to summarize
useful information from the history. We want a method that is
asymptotically consistent in the sense it will provably eventually
only choose between alternatives that satisfy an optimality property
related to the used criterion. We extend our work to the case where
there is side information that one can take advantage of and,
furthermore, we briefly discuss the active setting where an agent
takes actions to achieve desirable outcomes.
\def\contentsname{\centering\normalsize Contents}
{\parskip=-2.5ex\tableofcontents}
\end{abstract}


\newpage
\section{Introduction}\label{sec:Intro}

When studying long term sequence prediction one is interested in
answering questions like: What will the next $k$ observations be?
How often will a certain event or a sequence of events occur? What
is the average rate of a variable like cost or income? This can be
interesting for forecasting time series and for choosing policies
with desirable outcomes.

Hidden Markov Models \cite{CapMouRyd05,EphMer02} are often used for
long term forecasting and sequence prediction. In this article we
will restrict our study to models based on states that result from a
deterministic function of the history, in other words, states that
summarize useful information that has been observed so far. We will
consider finite state space maps with the property that given the
current state and the next observation we can determine the next
state. These maps are sometimes called Probabilistic-Deterministic
Finite Automata (PDFA) \cite{VidThoHig05a} and they have recently
been applied in reinforcement learning \cite{Mah10}. A particular
example of this is to use suffix trees \cite{Ris83,Sin96,McC96}.

Our goal is to prove consistency for our penalized Maximum
Likelihood criteria for picking a map from histories to states in
the sense that we want to eventually only choose between
alternatives that are optimal for prediction. The sense of
optimality could relate to predicting the next symbol, the next $k$
symbols or to have minimal entropy rate for an infinite horizon.

After the preliminary Section \ref{sec:Prelim} we begin our theory
development in Section \ref{sec:HtoS}. In our problem setting we
have a finite set $\Y$, a sequence $y_n$ of elements from $\Y$, and
we are interested in predicting the future of the sequence $y_n$. To
do this, being inspired by \cite{Hut09a} where general criteria for
choosing a feature map for reinforcement learning were discussed, we
first want to learn a \emph{feature map} $\Phi(y_{1:n})=s_n$ where
$y_{1:t}:=y_1,....,y_t$.

We would like the map to have the following properties:
\begin{enumerate}
\item The distribution for the sequence $s_n$ induced by the
distribution for the sequence $y_n$ should be that of a Markov chain
or should be a distribution which is indistinguishable from a Markov
chain for the purpose of predicting the sequence $y_n$.

\item We want as few states as possible so that we can learn a model
from a modest amount of data.

\item We want the model of the sequence $y_n$ that arises as a function of
the Markov chain $s_n$ to be as good as possible. Ideally it should
be the true distribution.
\end{enumerate}

Our approach consists of defining criteria that can be applied to
any class of $\Phi$, but later we restrict our study to a class of
maps that are defined by finite-state machines. These maps are
defined by introducing a deterministic function $\psi$ such that
$s_{n}=\psi(s_{n-1},y_n)$. If we have chosen such a map $\psi$ and a
first state $s_0$ then the sequence $y_n$ determines a unique
sequence $s_n$ and therefore we have also defined a map
$\Phi(y_{1:n})=s_n$.

In Section \ref{sec:Prelim} we provide some preliminaries on random
sequences and Hidden Markov Models. We introduce a class of ergodic
sequences which is the class of sequences that we work with in this
article. They are sequences with the property that an individual
sequence determines a distribution over infinite sequences.
We present our consistency theory by first presenting very generic
results in the beginning of Section \ref{sec:HtoS} and then we show
how various classes of maps and models fit into this. This has the
consequence that we first have results where we guarantee optimality
given that the individual sequence that we work with has certain
properties (and these results, therefore, have no ``almost sure'' in
the statement since the setting is not probabilistic) while in the
latter part we show that if we sample the sequence in certain ways
we will almost surely get a sequence with these properties.
In particular in Section \ref{sec:fsmmap} we will take a closer look
at suffix tree sources and maps based on finite state machines
related to probabilistic deterministic finite automata.
Section \ref{sec:Main} summarizes the findings in a main theorem
that says under some assumptions (a class of maps based on finite
state machines of bounded memory and ergodicity) we will recover the
true model (or the closest we can get to the true model).
Section \ref{sec:Side} contains a discussion of sequence prediction
with side information,
Section \ref{sec:Active} briefly discusses the active case where an
agent acts in an environment and earns rewards, and finally
Section \ref{sec:Conc} contains our conclusions.

\section{Preliminaries}\label{sec:Prelim}

In this section we will review some notions and results that the
rest of the article will rely upon. We start with random sequences
and then follows a section on Hidden Markov Models (HMM).

\paradot{Random Sequences}
Consider the set of all infinite sequences $y_t, t=1,2,...$ of
elements from a finite alphabet $\Y$. We equip the set with the
$\sigma$-algebra that is generated by the cylinder sets
$\Gamma_{y_{1:n}}=\{x_{1:\infty} |\ x_t=y_t, t=1,...,n\}$. A measure
with respect to this space is determined by its values on the
cylinder sets. Not every set of values is valid. We need to assume
that the measure of $\Gamma_{y_{1:t}}$ is the sum of the measures of
the sets $\Gamma_{y_{1:t}\tilde y}$ for all possible $\tilde y\in
\Y$. If we want it to be a probability measure we furthermore need
to assume that the measure of the whole space $\Y^\infty$ (which is
the cylinder  set  $\Gamma_\epstr$ of the empty string $\epstr$)
equals to one. The concept that is introduced in the following two
definitions is of central importance to this article. In particular
\emph{ergodic sequences} is the class of sequences that we intend to
model. They are sequences that can be used to define a distribution
over infinite sequences that we will be interested in learning.

\begin{definition}[Distribution defined from one sequence]
A sequence $y_{1:\infty}$ defines a probability distribution on
infinite sequences if the (relative) frequency of every finite
substring of $y_{1:\infty}$ converges asymptotically. The
probabilities of the cylinder sets are defined to equal those
limits:\\
\centerline{$
  \Gamma_{z_{1:m}}:=\lim_{n\to\infty}\#\{t\leq n:y_{t+1:t+m}=z_{1:m}\}/n
$}
\end{definition}

\begin{definition}[ergodic sequence]\label{def:erg}
We say that a sequence is ergodic if the frequencies of every finite
substring are converging asymptotically.
\end{definition}

As probabilistic models for random sequences we will in this article
focus on Hidden Markov Models (HMMs)  \cite{BauPet66,Pet69}. More
recent surveys on Hidden Markov Models are
\cite{EphMer02,CapMouRyd05}.

\paradot{Hidden Markov Models}
Here we define distributions over sequences of elements from a
finite set $\Y$ of size $Y$ based on an unobserved Markov chain of
elements from a finite state set $\S$ of size $S$.

\begin{definition}[Hidden Markov Model, HMM]
Assume that we have a Markov chain with an $S\times S$ transition
matrix $T=(T_{s,s'})$ and that we also have an $S\times Y$ emission
matrix $E=(E_{s,y})$ where $E_{s,y}$ is the probability that state
$s$ will generate outcome $y\in \Y$. If we introduce a starting
probability vector we have defined a probability distribution over
sequences of elements from $\Y$. This is called a Hidden Markov
Model (HMM).
\end{definition}

\paradot{Sequence Prediction}
One use of Hidden Markov Models (and functions of Markov chains) is
sequence prediction. Given a history $y_1,...,y_n$ we want to
predict the future $y_{n+1},...$. In some situations we know what
state we are in at time $n$ and that state then summarizes the
entire history without losing any useful information since the
future is conditionally independent of the past, given the current
state. If we are doing one step prediction we are interested in
knowing $Pr(y_{n+1}|s_n).$ We can also consider a zero step
lookahead (called filtering) $Pr(y_{n}|s_n)$ or an $m$ step
$Pr(y_{n+1},...,y_{n+m}|s_n).$ The $m$ step could also be just
$Pr(y_{n+m}|s_n).$ In a sense we can consider an infinite lookahead
ability evaluated by the entropy rate
$-\lim_{m\to\infty}\frac{1}{m}\log Pr(y_{n+1},...,y_{n+m}|s_n).$ If
the Markov chain is ergodic this limit does not depend on the state
$s_n$.

\paradot{Limit Theorems}
The following theory that is the foundation for studying consistency
of HMMs was developed in \cite{BauPet66} and \cite{Pet69}. See
\cite{CapMouRyd05} chapter 12 for the modern state of the art.

\begin{definition}[ergodic Markov chain]
A Markov chain (and the stochastic matrix that contains its
transition probabilities) is called ergodic if it is possible to
move from state $s$ to state $s'$ in a finite number of steps
for all $s$ and $s'$.
\end{definition}

The following theorem \cite{CapMouRyd05} introduces the generalized
cross-entropy $H$ and shows that it is well defined and that it can
be estimated for ergodic HMMs. It can be interpreted as the
(idealized) expected number of bits needed for coding a symbol
generated by a distribution defined by $\t_0$ but using the
distribution defined by $\t$.

\begin{theorem}[ergodic HMMs]\label{thm:erg}
If $\t$ and $\t_0$ are HMM parameters where the transition
matrix for $\t_0$ is an ergodic stochastic matrix, then there
exists a finite number $H(\t_0,\t)$ (which can also be
defined as $\lim_{n\to\infty} H_{n,s}(\t_0,\t)$ for any
initial state $s$ where $H_{n,s}(\t_0,\t):=
\frac{1}{n}\EE_{\t_0}\log{Pr(y_1,...,y_n|s_0=s,\t)})$ such
that $P_{\t_0}$ a.s.
\beqn
  -\lim_{n\to\infty}\frac{1}{n}\log{Pr(y_1,...,y_n|\ \t)} ~=~ H(\t_0,\t)
\eeqn
and the convergence is uniform in the parameter space.
\end{theorem}

\begin{definition}[Equivalent HMMs]
For an HMM $\t_0$, let $M[\t_0]$ be the set of all $\t$
such that the HMM with parameters $\t$ define the same
distribution over outcomes as the HMM with parameters $\t_0$.
\end{definition}

\begin{theorem}[Minimal cross-entropy for the truth and only the truth]
$H(\t_0,\t)\geq H(\t_0,\t_0)$ with equality if and
only if $\t\in M[\t_0]$.
\end{theorem}

\section{Maps From Histories To States}\label{sec:HtoS}

Given a sequence of elements $y_n$ from a finite alphabet we want to
define a map $\Phi:\Y^*\to\S$, which maps histories (finite strings)
of elements to states $\Phi(y_{1:n})=s_n$. The reasons for this
include, as was explained in the introduction, in particular the
ability to learn a model efficiently. Suppose that every $\Phi$
under consideration is such that the size of its state space $\S$ is
a finite number that depends on $\Phi$.

We are also interested in the case when we have side information
$x_n\in \X$ and we define a map $\Phi:(\X\times\Y)^*\to\S$. In this
more general case the models that we consider for the sequence $y$
will have hidden states while in the case without side information
the state (given the $y$ sequence) is not hidden. We have two
reasons for expressing everything in an HMM framework.  We can model
long-range dependence in the $y_n$ sequence through having states
and we include the more general case where there is side
information.

\begin{definition}[Feature sequence/process]
A map $\Phi$ from finite strings of elements from $\Y$ (or $\X\times\Y$) to elements
in a finite set $\S$ and a sequence $y_{1:n}$ induces a state
sequence $s_{1:n}$. Define an HMM through maximum likelihood
estimation: The sequence $s_t=\Phi(y_{1:t})$ gives transition
matrix $T(n)=(T_{s,s'})$ of probabilities
\beqn
  T_{s,s'}(n) ~:=~ \frac{\#\{t\leq n|s_t=s,\ s_{t+1}=s'\}}{\#\{t\leq n|s_t=s\}}
\eeqn
and emission matrix $E(n)$ of probabilities
\beqn
  E_{s,y}(n) ~:=~ \frac{\#\{t\leq n|s_t=s,\ y_{t}=y\}}{\#\{t\leq n| s_t=s\}}.
\eeqn
Denote those HMMs by $\hat{\t}_n:=(T(n),E(n))$. We will refer to the
sequence $\hat{\t}_n$ as the parameters corresponding to $\Phi$ or
generated by $\Phi$.
\end{definition}

We will first state results based on some
generic properties that we have defined with just the goal of making
the proofs work. Then we will show that some more easily understandable
cases will satisfy these properties. We structure it this way not
only for generality but also to make the proof techniques clearer.

\paradot{Ergodic Sequences}\label{sec:ergseq}
We begin by defining the fundamental ergodicity properties that we
will rely upon. We provide asymptotic results for individual
sequences that satisfy these properties. In the next two subsections
we identify situations where we will almost surely get such a
sequence which satisfies these ergodicity properties.

\begin{definition}[ergodic w.r.t.\ $\Phi$]\label{def:phierg}
As stated in Definition \ref{def:erg}, we say that a sequence $y_t$
is ergodic if all substring frequencies converge as $n\to\infty$.
Furthermore we say that
\\1. the sequence $y_t$ is ergodic with respect to a
map $\Phi(y_{1:t})=s_t$ if all state transition frequencies
$T_{s,s'}(n)$ and emission frequencies $E_{s,y}(n)$ converge as
$n\to\infty$.
\\2. the sequence $y_t$ is ergodic with respect to a
class of maps if it is ergodic with respect to every map in the
class.
\end{definition}

\begin{definition}[HMM-ergodic]\label{def:hmmerg}
We say that a sequence $y_t$ is HMM-ergodic for a set of HMMs
$\Theta$ if there is an HMM with parameters $\t_0$ such that
\beqn
  -\frac{1}{n}\log Pr(y_1,...,y_n\ |\ \t) ~\to~ H(\t_0,\t)
\eeqn
uniformly on compact subsets of $\Theta$.
\end{definition}

\begin{definition}[Log-likelihood]
$L_n(\Phi)=-\log{Pr(y_1,...,y_n|\hat{\t}_n)}$
\end{definition}

We will prove our consistency results by first proving consistency
using Maximum Likelihood (ML) for a finite class of maps and then we
prove that we can add a sublinearly growing model complexity penalty
and still have consistency.

\begin{proposition}[HMM consistency of ML for finite class]\label{genfinite}
Suppose that $y_t$ is HMM-ergodic for the parameter set $\Theta$
with optimal parameters (in the sense of Definition
\ref{def:hmmerg}) $\t_0$, $y_t$ is ergodic for the finite class of
maps $\{\Phi_i\}_{i=1}^K$ and suppose that $\t_i\in\Theta$ are the
limiting parameters generated by $\Phi_i$. Then it follows that
there is $N<\infty$ such that for all $n\geq N$  the map $\Phi_i$
selected by minimizing $L_n$ generates parameters $\hat{\t}_i^n$
whose limit is in $\arg\min_{\t_i} H(\t_0,\t_i)$.
\end{proposition}

\begin{proof}
It follows from Definition \ref{def:hmmerg} and continuity (in $\t$)
of the log-likelihood that
\beqn
  \lim_{n\to\infty} \frac{1}{n}L_n(\Phi_i) ~=~ H(\t_0,\t_i)
\eeqn
since the convergence in Definition \ref{def:hmmerg} is uniform.
Note that the parameters that define the log-likelihood
$L_n(\Phi_i)$ can be different for every $n$ so the uniformity of
the convergence is needed to draw the conclusion above. By
Definition \ref{def:phierg} we know that if $\hat{\t}_n^i$ are the
parameters generated by $\Phi_i$ at time $n$, then
$\lim_{n\to\infty}\hat{\t}_n^i=\t_i$ exists for all $i$. It follows
that if $\t_i\notin \arg\min_{\t_j} H(\t_0,\t_j)$ then there must be
an $N<\infty$ such that $\Phi_i$ is not selected at times $n\geq N$.
Since there are only finitely many maps in the class there will be a
finite such $N$ that works for all relevant $i$. \qed\end{proof}

\begin{definition}[HMM Cost function]\label{def:hmmcost}
If the HMM with parameters $\hat{\t}_n$ that has been estimated
from $\Phi$ at time $n$ has $S$ states, then let
\beqn
  \Cost_n(\Phi) ~=~ -\log{Pr(y_1,...,y_n|\hat{\t}_n)}+pen(n,S)
\eeqn
where $pen(n,S)$ is a positive function that is increasing in both
$n$ and $S$ and is such that $pen(n,S)/n\to 0$ for $n\to\infty$
for all $S$.
\end{definition}

We call the negative log-probability term the \emph{data coding
cost} and the other term is the \emph{model complexity penalty}.
They are both motivated by coding (coding the data and the model).
For instance in MDL/MML/BIC, $pen(n,S)={d\over 2}\log n+O(1)$, where
$d$ is the dimensionality of the model $\t$.

\begin{proposition}\label{genfewer}
Suppose that $\Phi_0$ has optimal limiting parameters $\t_0$ with as
few states as possible. In other words if an HMM has fewer states
than the HMM defined by $\t_0$, then it has a strictly larger
entropy rate. We use a (finite, countable, or uncountable) class of
maps that includes only $\Phi_0$ and maps that have strictly fewer
states. We assume that all the maps generate converging parameters.
Then there is an $N$ such that the function $\Cost$ is minimized by
$\Phi_0$ at all times $n\geq N$.
\end{proposition}

\begin{proof}
Suppose that $\t_0$ has $S_0$ states. We will use a bound for how
close one can get to the true HMM using fewer states. We would like
to have a constant $\eps>0$ such that $H(\t_0,\t)>H(\t_0,\t_0)+\eps$
for all $\t$ with fewer then $S_0$ states. The existence of such an
$\eps$ follows from continuity of $H$ (which is actually also
differentiable \cite{BauPet66}), the fact that the HMMs with fewer
than $S_0$ states can be compactly (in the parameter space) embedded
into the space of HMMs with exactly $S_0$ states, and that this
embedded subspace has a strictly positive minimum Euclidean distance
from $\t_0$ in this parameter space.

The existence of $\eps>0$ with this property implies the existence
of $D>0$ such that the alternatives with fewer than $S_0$ states
have, for large $n$, at least $Dn$ worse log probabilities than the
distribution $\t_0$. Therefore the penalty term (for which
$pen(n,S)/n\to 0$) will not be able to indefinitely compensate for
the inferior modeling.
\qed\end{proof}

\begin{theorem}[HMM consistency of Cost for finite class]\label{corgenfinite}
Proposition \ref{genfinite} is also true for $\Cost$.
\end{theorem}

\begin{proof}
$H(\t_0,\t_k)<H(\t_0,\t_j)$ implies that there is a constant $C>0$
such that for large $n$, $L_n(\Phi_j)-L_n(\Phi_k)\geq Cn$. Since
$pen(n,S)/n\to 0$ for $n\to\infty$ we know that any difference
in model penalty will be
overtaken by the linearly growing difference in data code length.
\qed\end{proof}

\paradot{Maps that induce HMMs}\label{sec:indhmm}
In this section we will assume that we use a class of maps whose
states we know form a Markov chain.

\begin{definition}[Feature Markov Process, $\mathbf\Phi$MP]
Suppose that
\beqn
 Pr(y_{n}|\Phi_0(y_1),...,\Phi_0(y_{1:n})) ~=~ Pr(y_n|\Phi_0(y_{1:n}))
\eeqn
and that the state sequence is Markov, i.e.
\beqn
  Pr(\Phi_0(y_{1:n})|\Phi_0(y_1),...,\Phi_0(y_{1:n-1})) ~=~ Pr(\Phi_0(y_{1:n})|\Phi_0(y_{1:n-1})).
\eeqn
Then we say that $\Phi_0$ induces an HMM. We call HMMs induced by
$\Phi_0$, Feature Markov Process ($\Phi$MP). If the HMM that
is defined this way by $\Phi_0$ is the true distribution for the
sequence $y_1,y_2,...$, then we say that ``$\Phi_0$ is correct''.

We will only discuss the situation when the true HMM is ergodic so
we will only say that there is a correct $\Phi_0$ in those
situations, hence the statement $\Phi_0$ is correct will contain the
assumption that the truth is ergodic.
\end{definition}

\begin{example}
The map $\Phi$ which sends everything to the same state always
induces an HMM but, unless the sequence $y_1,y_2,...$ is i.i.d, it
is not correct.
\eoe\end{example}

\begin{proposition}[Convergence of estimated distributions]
If $\Phi_0$ is correct then $P_{\hat{\t}_n}\to P_{\t_0}$ for
$n\to\infty$ (as distributions on finite strings of a (any) fixed
length), where $P_{\t_0}$ is the true HMM distribution for the
outcomes, $P_\t$ is the HMM distribution defined by $\t$ and
$\hat{\t_n}$ are the parameters generated by $\Phi_0$.
\end{proposition}

\begin{proof}
We are estimating the parameters $\hat{\t}_n$ through maximum
likelihood for the generated sequence of states. Consistency of
maximum likelihood estimation for Markov chains implies that
$\hat{\t}_n\to\t_0$. This implies the proposition due to
continuity with respect to the parameters of the likelihood (for any
finite sequence length).
\qed\end{proof}

\begin{proposition}[Inducing HMM implies drawing ergodic sequences]
If we have a set of maps that induce HMMs and the sequence $y_t$ is
drawn from one of the induced ergodic HMMs, then almost surely \\1.
$y_t$ is HMM-ergodic\\2. we will draw an ergodic sequence
$y_t$ with respect to the considered class of maps.
\end{proposition}

\begin{proof}
1. is a consequence of Theorem \ref{thm:erg}.
\\2. This follows from consistency of
maximum likelihood for Markov chains (generalized law of large
numbers) since the claim is that state transition frequencies and
emission frequencies converge.
\qed\end{proof}

\section{Maps based on Finite State Machines (FSMs)}\label{sec:fsmmap}

We will in this section consider maps of a special form that are
related to PDFAs. We will assume that $\Phi$ is such that there is a
$\psi$ such that
\beqn
  \Phi(y_{1:n}) ~=~ \psi(\Phi(y_{1:n-1}),y_n).
\eeqn
In other words, the current state is derived deterministically from
the previous state and the current perception. Given an initial
state the state sequence is then deterministically determined by the
perceptions and therefore the combination  of $\psi$ with an initial
state defines a map $\Phi$ from histories to states. This class of
maps $\Phi$ can also define a class of probabilistic models of the
sequence $y_n$ by assuming that $y_n$ only depends on
$s_{n-1}=\Phi(y_{1:n-1})$.  This leads to the formula
\beqn
  Pr(s'|s) ~=~ \sum_{y:\psi(s,y)=s'} Pr(y|s)
\eeqn
and as a result we have defined an HMM for the sequence $y_n$.

\begin{definition}[Sampling from FSM]\label{def:samplemap}
If we follow the procedure above we say that we have sampled the
sequence $y_t$ from the FSM. If the Markov chain of states is
ergodic we say that we have sampled $y_t$ ergodically from the FSM.
\end{definition}

\paradot{Suffix Trees}
We consider a class of maps based on FSMs that can be expressed
using Suffix Trees \cite{Ris86} with the same states (suffixes) as
the FSM. The resulting models are sometimes called FSMX sources. A
suffix tree is defined by a suffix set which is a set of finite
strings. The set must have the property that none of the strings is
an ending substring (a suffix) of another string in the set and such
that any sufficiently long string ends with a substring in the
suffix set. Given any sufficiently long string we then know that it
ends with exactly one of the suffixes from the suffix set. If the
suffix set furthermore has the property that given the previous
suffix and the new symbol there is exactly one element (state) from
the suffix set that can (and is) the end of the new longer string,
then it is an FSMX source. Another terminology says that the suffix
set is FSM closed. The property implies (directly by definition)
that there is a map $\psi$ such that $\psi(s_{t-1},y_t)=s_t$.

The following proposition shows a very nice connection between
ergodic sequences and FSMX sources which will be generalized in
Proposition \ref{FSMerg} to more general sources based on
bounded-memory FSMs.

\begin{proposition}[ergodicity of suffix trees]\label{suffix}
If we have a set of maps based on FSMs that can be expressed by
suffix trees, and the sequence $y_t$ is sampled ergodically
(Definition \ref{def:samplemap}) using one of the maps, then almost
surely we get a sequence $y_t$ that is ergodic with respect to the
considered class of maps and $y_t$ is HMM-ergodic.
\end{proposition}

\begin{lemma}\label{sufone}
If the sequence $y_t$ is ergodic, then the state transition
frequencies and emission (of $y$) frequencies for a FSM
closed suffix tree are converging.
\end{lemma}

\begin{proof}
Let the map $\Phi$ be defined by the suffix set in question. Suppose
that $s'$ is a suffix that can follow directly after $s$. This means
that there is a symbol $y$ such that if you concatenate it to the
end of the string $s$, then this new string $\tilde{s}$ ends with
the string $s'$. This means that whenever a string of symbols
$y_{1:n}$ ends with $\tilde{s}$, then the sequence of
states generated by applying the map $\Phi$ to the sequence
$y_{1:n}$ will end with $s_{n-1}=s$ and $s_n=s'$.  It is
also true that whenever the state sequence ends with $ss'$ then
$y_{1:n}$ ends with $\tilde{s}$. Therefore, the counts (of $ss'$ in
the state sequence and $\tilde{s}$ in the $y$ sequence) up until any
finite time point are also equal. We will in this proof say that
$\tilde{s}$ is the string that corresponds to $ss'$.
\\ Given any ordered pair of states $(s,s')$
where $s'$ can follow $s$, let $c_{s,s'}(n)$ be the number of times
$ss'$ occurs in the state sequence up to time $n$ and let
$d_{s,s'}(n)$ be the number of times the string $\tilde{s}$ that
corresponds to $ss'$ has occurred. We know that
$c_{s,s'}(n)=d_{s,s'}(n)$ for any such pair $ss'$ and any $n$. If
$s'$ cannot follow $s$ we let both $c_{s,s'}=0$ and $d_{s,s'}=0$.
The state transition frequency for the transition from $s$ to $s'$
up until time $n$ is
\beqn
  \frac{c_{s,s'}(n)}{\sum_{s'} c_{s,s'}(n)}
  ~=~ \frac{d_{s,s'}(n)}{\sum_{s'} d_{s,s'}(n)}
  ~=~ \frac{d_{s,s'}(n)}{d_{s}(n)}
  ~=~ \frac{d_{s,s'}(n)}{n}\frac{n}{d_{s}(n)}
\eeqn
where $d_s(n)$ is the number of times that the string that defines
$s$ has occurred up until time $n$ in the $y$ sequence. The right
hand side converges to the frequency of the string $\tilde{s}$
divided by the frequency of the string that defines $s$.
Thus we have proved that state transition frequencies converge.
Emissions work the same way.
\qed\end{proof}

\begin{lemma}\label{suftwo}
If we sample $y_t$ ergodically from a suffix tree FSM, then the
frequency for each finite substring will converge almost surely. In
other words the sequence $y_t$ is almost surely ergodic.
\end{lemma}

\begin{proof}
If the suffix tree defines an FSM as we have defined it above, the
states of the suffix tree will form an ergodic Markov chain. An
ergodic Markov chain is stationary. For any state and finite string
of perceptions there is a certain fixed probability of drawing the
string in question. The frequency of the string $str$ is $\sum_s
Pr(s)Pr(str|s)$ where $Pr(s)$ is the stationary probability of
seeing $s$ and $Pr(str|s)$ is the probability of directly seeing
exactly $str$ conditioned on being in state $s$. It follows from the
law of large numbers that the frequency of any finite string $str$
converges.

Another way of understanding this result is that it is implied by
the convergence of the frequency of any finite string of states in
the state sequence.
\qed\end{proof}

\begin{proof} {\bf of Proposition \ref{suffix}.}
Lemma \ref{sufone} and Lemma \ref{suftwo} together imply the
proposition since they say that if we sample from a suffix tree then
we almost surely get converging frequencies for all finite
substrings and this implies converging transition frequencies for
the states from any suffix tree.
\qed\end{proof}

\paradot{Bounded-Memory FSMs}
We here notice that the reasons that the suffix tree theory above
worked actually relate to a larger class, namely a class of FSMs
where the internal state is determined by at most a finite number of
previous time steps in the history.

\begin{definition}[bounded memory FSM]
Suppose that there is a constant $\kappa$ such that if we know the
last $\kappa+1$ perceptions $y_{t-\kappa},...,y_t$ then the present
state $s_t$ is uniquely determined. Then we say that the FSM has
memory of at most length $\kappa$ (not counting the current) and
that it has bounded memory.
\end{definition}

\begin{proposition}[ergodicity of FSMs]\label{FSMerg}
1. Consider a sequence $y_t$ whose finite substring frequencies
converge (i.e.\ the sequence is ergodic) and an FSM of bounded
memory, then the sequence is ergodic with respect to the map defined
by the FSM.
\\ 2. If we sample a sequence $y_t$ ergodically from an FSM with bounded
memory then almost surely $y_t$ is HMM-ergodic and its finite
substring frequencies converge.
\end{proposition}

\begin{proof}
The proof works the same way as for suffix tree FSMs. If an FSM has
finite memory of length $\kappa$ then there is a suffix tree of that
depth with every suffix of full length and every state of the FSM is
a subset of the states of that suffix tree. The FSM is a partition
of the suffix set into disjoint subsets. Every state transition for
the FSM is exactly one of a set of state transitions for the suffix
tree states and the frequency of every ordered pair of suffix tree
states converge almost surely as before. Therefore, the state
transition frequencies for the FSM will almost surely converge.

A distribution that is defined using an FSM of bounded memory can
also be defined using a suffix tree, so 2. reduces to this case
\qed\end{proof}

\section{The Main Result For Sequence Prediction}\label{sec:Main}

In this section we summarize our results in a main theorem. It
follows directly from a combination of results in previous sections.
They are stated with respect to our main class of maps, namely the
class that is defined by bounded-memory FSMs. The generating models
that we consider are models that are defined from a map in this
class in such a way that the states form an ergodic Markov chain. We
refer to this as sampling ergodically from the FSM. Our conclusion
is that we will under these circumstances eventually only choose
between maps which generate the best possible HMM parameters that
can be achieved for the purpose of long-term sequence prediction.
The model penalty term will influence the choice between these
options towards simpler models.

The following theorem guarantees that we will almost surely
asymptotically find a correct HMM for the sequence of interest under
the assumption that it is possible.

\begin{theorem}\label{thm:passive}
If we consider a finite class of maps $\Phi_i, i=0,1,...,k$ based on
finite state machines of bounded memory and if we sample ergodically
from a finite state machine of bounded memory, then there almost
surely exist limiting parameters $\t_i$ for all $i$ and there is
$N<\infty$ such that for all $n\geq N$  the map $\Phi_i$ selected at
time $n\geq N$ by minimizing $\Cost$, generates parameters whose
limit is $\t_0$ which is assumed to be the optimal HMM parameters.
\end{theorem}

\begin{proof}
We are going to make use of Proposition \ref{FSMerg} together with
Theorem \ref{corgenfinite}. Proposition \ref{FSMerg} shows that our
assumptions imply the assumptions of Theorem \ref{corgenfinite} which
provides our conclusion.
\qed\end{proof}

\paradot{Extension to countable classes}
To extend our results from finite to countable classes of maps we
need the model complexity penalty to be sufficiently rapidly growing
in $n$ and $m$. This is also necessary if we want to be sure that we
eventually find a minimal representation of the optimal model that can
be achieved by the class of maps.

\begin{proposition}[Consistency for countable class]\label{countable}
Suppose that we have a countable class of maps $\Phi_i,\ i=0,1,...$ and
\begin{enumerate}
\item Suppose that our class is such that for every finite $k$,
    there are at most finitely many maps with at most $k$
    states.
\item Suppose that $\theta_0$ is an optimal HMM for the sequence
    $y_t$, that it has $m$ states and that $\theta_0$ is the
    limit of the parameters generated by $\Phi_0$. Furthermore,
    suppose that there is finite $N$ such that whenever $n>N$,
    $\tilde{m}>m$ and $\tilde{\theta}$ is any HMM with
    $\tilde{m}$ states we have $pen(n,m)-\log
    P_{\hat{\theta}_0^n}(y_1,...,y_n)<
    pen(n,\tilde{m})-\log{P_{\tilde{\theta}}(y_1,...,y_n)}$.
    where $\hat{\theta}_0^n$ are the parameters generated by
    $\Phi_0$.
\end{enumerate}
then Theorem \ref{corgenfinite} is true  also for this countable
class and we will furthermore eventually pick a map with at most $m$
states.
\end{proposition}

\begin{proof}
The idea of the proof is to reduce the countable case to the finite
case that we have already proven by using that when $n>N$ we will
never pick a $\Phi$ with more than $m$ states and then use the first
property to say that the remaining class if finite. This reduction
also shows that we will eventually not pick a map with more states
than $m$. \qed
\end{proof}

The first property in the proposition above holds for the class of
suffix trees and for the class based on FSMs with bounded memory.
The second property, but with the HMM maximum likelihood parameters
$\theta(n)$ with $m$ states (while we have ML for a sequence of
states and observations)  will almost surely hold if the penalty is
such that we have strong consistency for the HMM criteria
$\theta^*=\arg\max \log{P_\theta(y_1,...,y_n)}-pen(n,m)$. This is
studied in many articles, e.g. \cite{GasBou03} where strong
consistency is proven for a penalty of the form $\beta(m)\log{n}$
where $\beta$ is a cubic polynomial. Note that in the case without
side information (if our map has the properties that
$\Phi_0(y_{1:n})$ determine $y_n$ and that $\Phi(y_{n-1})$ and $y_n$
determine $\Phi(y_{1:n})$) the emissions are deterministic and the
state sequence generated by any map is determined by the $y$
sequence. This puts us in a simpler situation akin to the Markov
order estimation problem \cite{Fin96, Csi00} where it is studied
which penalties (e.g.\ BIC) will give us property 2. above.

\begin{conjecture}
We almost surely have Property 2. from Proposition \ref{countable}
for the BIC penalty studied in \cite{Csi00}.
\end{conjecture}

\section{Sequence Prediction With Side Information}\label{sec:Side}

In this section we will broaden our problem to the setting where we
have side information available to help in our prediction task. In
our problem setting we have two finite sets $\X$ and $\Y$, a
sequence $p_n=(x_n,y_n)$ of elements from $\X\times \Y$, and we are
interested in predicting the future of the sequence $y_n$. To do
this we first want to learn a {\emph {feature map}}
$\Phi(p_{1:n})=s_n$. In other words we want our current state to
summarize all useful information from both the $x$ and $y$ sequence
for the purpose of predicting the future of $y$ only.

One obvious approach is to predict the future of the entire sequence
$p$, i.e. predicting both $x$ and $y$ and then in the end only
notice what we find out about $y$. This brings us back to the case we
have studied already, since from this point of view there is no side
information. A drawback with that approach can be that we create an
unnecessarily complicated state representation since we are really
only interested in predicting the $y$ sequence.

In the case when there is no side information, $s_t=\Phi(y_{1:t})$.
An important difference of the case with side information is that
the sequence $s_{1:t}$ depends on both $y_{1:t}$ and $x_{1:t}$.
Therefore for the latter case, if we would like to consider a
distribution for $y$ only, $y_1,...,y_n$ does not determine the
state sequence $s_1,...,s_n$:
\beqn
  Pr(y_1,...,y_n|\hat{\t}_n) ~= \sum_{s_{1:n},x_{1:n}}
  Pr(s_1,...,s_n)Pr(x_1,...,x_n,y_1,...,y_n|s_1,...,s_n,\hat{\t}_n).
\eeqn
This is expression is of course also true in the absence of side
information $x$, but then the sum collapses to one term since there is
only one sequence of states $s_{1:n}$ that is compatible with
$y_{1:n}$.

An alternative to using the $\Cost$ criteria on the $p$ sequence is
to only model the $y$ sequence and let
\beqn
  L_n(\Phi) ~=~ -\log{Pr(y_1,...,y_n|\hat{\t}_n)}
\eeqn
and then define $\Cost$ in exactly the same way as before. This cost
function was called $\ICost$ in \cite{Hut09a}.

\begin{theorem}\label{thm:icost}
Theorem \ref{thm:passive} is true for sequence prediction with side
information using
\beqn
  \ICost_n(\Phi_i) ~=~ -\log{Pr(y_1,...,y_n|\hat{\t}_n)}+pen(n,S)
\eeqn
if we define ``sample ergodically'' to refer to the sequence
$p_t=(x_t,y_t)$ instead of $y_t$.
\end{theorem}

\begin{proof}
The proofs work exactly as they are written for the case without
side information.
\qed\end{proof}

Note that a map that is optimal for predicting the $y$ sequence can
have fewer states than a minimal map that can generate the model of
the $p$ sequence.

It is interesting to note that the interpretation of this result is
not as clear as the case without side information. It guarantees
that, given enough history, the chosen $\Phi$ can and will (with the
asymptotic parameters) define the correct model for the $y_t$
sequence but the $x_t$ sequence has only played a part in the
estimation and we are not guaranteed that we will make use of the
extra information if it does not impact the entropy rate. In
particular it is true if the information in $x_t$ is only helpful
for a finite number of time steps forward. In this case that gain
will not affect the entropy rate which is a limit of averages. We
have a more conclusive result for the case with side information when we
use the first mentioned approach of applying $\Cost$ to the sequence
$p$, since we proved consistency in the previous section in the sense
of finding the true model when possible.

If we have injective maps $\Phi$, e.g.\ maps defined by non-empty
suffix trees, then we can rewrite $\Cost$ in a form that was used in
\cite{Hut09a} also more generally. Therein a cost called
\emph{original cost} was defined as follows:
\begin{definition}[OCost]
\beqn
  \OCost ~=~ -\log{Pr(s_1,...,s_n)}-\log{Pr(y_1,...,y_n|s_1,...,s_n,\hat{\t}_n)}+pen(n,S).
\eeqn
\end{definition}
\begin{remark}
If $\Phi_i$ is injective and we calculate $\Cost$ in the side
information case then $\Cost~=~ \OCost$.
\eor\end{remark}

If we have no side information both $\OCost$ and $\ICost$ will be the
same as $\Cost$ but they may differ when there is side information
available. We remarked above that if we consider only injective $\Phi$
(e.g. non-empty suffix tree based maps) then $\OCost$ equals using
$\Cost$ on the joint sequence $p_t=(x_t,y_t)$.  As noted in
\cite{Hut09a} $\OCost$ penalizes having many states more than
$\ICost$ does and when considering non-injective $\Phi$ one risks
getting a smaller than desired state space.

\section{The Active Case}\label{sec:Active}

In this very brief section we will discuss how to map the active
case to the previously introduced notions. The active case will be
treated in depth in future articles. In the active case
\cite{RusNor10,SutBar98} we have an agent that interacts with an
environment. The agent perceives observations $o_t$ and real-valued
rewards $r_t$ and the agent takes actions $a_t$ from a finite set of
possible actions $\cal{A}$ with the goal of receiving high total
reward in some sense. We will denote the events that have just
occurred when the agent will take an action at time step $t$, i.e.\
$a_{t}$, $o_t$, and $r_t$ by $e_t$. We consider maps based on FSMs
(PDFAs) that takes event sequences $e_t$ as input. In the previous
section's notation $x_t=(o_t,a_t)$ and $y_t=r_t$ and $p_t=e_t$. We
chose this since we are interested in predicting which future
rewards will result from actions chosen with the help of the
observations. This would give us the possibility of determining
which actions will earn the highest rewards.

At time $t-1$ the past $e_1,...,e_{t-1}$ determines $s_{t-1}$ and
the agent takes an action $a_{t-1}$ and $o_{t}$ and $r_{t}$ are
generated according to distributions that only depend on $s_{t-1}$
and $a_{t-1}$. Then we have generated $e_{t}$ and
$s_{t}=\psi(s_{t-1},e_{t})$.

\begin{definition}\label{def:ergenv}
The above describes what we mean when we say that the FSM generates
the environment. We say that the FSM generates the environment
ergodically, if for any sequence of actions chosen such that the
action frequencies for any state converge asymptotically, we will
have state transitions and emission frequencies that converge almost
surely to an ergodic HMM.
\end{definition}

\begin{proposition}\label{FSMpr}
Suppose that we have an FSM of bounded-memory generating the
environment ergodically and the action frequencies for any state
converge asymptotically, then we will almost surely generate an
ergodic sequence of events and the reward sequence is HMM-ergodic.
\end{proposition}

\begin{proof}
The situation reduces through Definition \ref{def:ergenv} to that of
Proposition \ref{FSMerg}.
\qed\end{proof}

\begin{theorem}\label{thm:active}
If we consider a finite class of maps $\Phi_i,  i=0,1,...,k$ based
on finite state machines of bounded memory and if the environment is
generated ergodically by a finite state machine of bounded memory and
if the action frequencies for any internal state of the generating
finite state machine converge, then there almost surely exist
limiting state transition parameters $\t_i$ for all $i$ and
there is $N<\infty$ such that for all $n\geq N$  the map $\Phi_i$
selected by minimizing $\ICost$ at time $n\geq N$ generates
parameters whose limit is $\t_0$ which is the optimal HMM.
\end{theorem}

\begin{proof}
We combine Proposition \ref{FSMpr} with Theorem \ref{thm:icost}.
\qed\end{proof}

How to choose the actions to make the implications for reinforcement
learning what we want them to be is the subject of ongoing work
\cite{Hut09a}.

\section{Conclusions}\label{sec:Conc}

Feature Markov Decision Processes were introduced \cite{Hut09a} as a
framework for creating generic reinforcement learning agents that
can learn to perform well in a large variety of complex
environments. It was introduced as a concept without theory or
empirical studies. First empirical results are reported in
\cite{Mah10}. Here we provide a consistency theory by focusing on
the sequence prediction case with and without side information.
We briefly discuss the active case where an agent takes actions that
may affect the environment. The active case and empirical studies is
the subject of ongoing and future work.

\paradot{Acknowledgement}
We thank the reviewers for their meticulous reading and valuable
feedback and the Australian Research Council for support under grant
DP0988049.


\begin{small}
\newcommand{\etalchar}[1]{$^{#1}$}

\end{small}

\end{document}